# Comparator Networks

Weidi Xie, Li Shen and Andrew Zisserman

Visual Geometry Group, Department of Engineering Science
University of Oxford
{weidi,lishen,az}@robots.ox.ac.uk

**Abstract.** The objective of this work is set-based verification, e.g. to decide if two sets of images of a face are of the same person or not. The traditional approach to this problem is to learn to generate a feature vector per image, aggregate them into *one* vector to represent the set, and then compute the cosine similarity between sets. Instead, we design a neural network architecture that can directly learn set-wise verification. Our contributions are: (i) We propose a Deep Comparator Network (DCN) that can ingest a pair of sets (each may contain a *variable* number of images) as inputs, and compute a similarity between the pair – this involves attending to multiple discriminative local regions (landmarks), and comparing local descriptors between pairs of faces; (ii) To encourage high-quality representations for each set, internal competition is introduced for recalibration based on the landmark score; (iii) Inspired by image retrieval, a novel hard sample mining regime is proposed to control the sampling process, such that the DCN is complementary to the standard image classification models. Evaluations on the IARPA Janus face recognition benchmarks show that the comparator networks outperform the previous state-of-the-art results by a large margin.

## 1 Introduction

The objective of this paper is to determine if two sets of images are of the same object or not. For example, in the case of face verification, the set could be images of a face; and in the case of person re-identification, the set could be images of the entire person. In both cases the objective is to determine if the sets show the same person or not.

In the following, we will use the example of sets of faces, which are usually referred to as 'templates' in the face recognition literature, and we will use this term from here on. A template could consist of multiple samples of the same person (e.g. still images, or frames from a video of the person, or a mixture of both). With the great success of deep learning for image classification [1–4], by far the most common approach to template-based face verification is to generate a vector representing each face using a deep convolutional neural network (CNN), and simply average these vectors to obtain a vector representation for the entire template [5–8]. Verification then proceeds by comparing the template vectors with some similarity metrics, e.g. cosine similarity. Rather than improve on this simple combination rule, the research drives until now has been to improve the performance of the single image representation by more sophisticated training losses, such as Triplet Loss, PDDM, and Histogram Loss [6, 7, 9–12]. This approach has achieved very impressive results on the challenging benchmarks, such as the IARPA IJB-B and IJB-C datasets [13, 14].

However, this procedure of first generating a single vector per face, and then simply averaging these, misses out on potentially using more available information in four ways:

First, *viewpoint conditioned similarity* – it is easier to determine if two faces are of the same person or not when they have a similar pose and lighting. For example, if both are frontal or both in profile, then point to point comparison is possible, whereas it isn't if one is in profile and the other frontal;

Second, *local landmark comparison* – to solve the fine-grained problem, it is essential to compare discriminative congruent 'parts' (local regions of the face), such as, an eye with an eye, or a nose with a nose.

Third, *within template weighting* – not all images in a template are of equal importance, the features derived from a low resolution or blurred face is probably of less importance than the ones coming from a high-resolution perfectly focussed face;

Fourth, *between template weighting* – what is useful for verification depends on what is in both templates. For example, if one template has only profile faces, and the second is all frontal apart from one profile instance, then it is likely that the single profile instance in the second template is of more importance than the frontal ones.

The simple average combination rule cannot take advantage of any of these four, for example, unweighted average pooling ignores the difference in the amount of information provided by each face image, and an aberrant image, such as one that is quite blurred, can have a significant effect (since most blurred face images look similar).

In this paper, we introduce a *Deep Comparator Network* (DCN), a network architecture designed to compare pairs of templates (where each template can contain an arbitrary number of images). The model consists of three modules: *Detect, Attend* and *Compare*, as illustrated in Figure 1, that address the four requirements above: in the *Detect* module, besides the dense feature representation maps, multiple discriminative landmark detectors act on each input image and generate the score maps; the *Attend* module normalizes the landmark responses over the images within template, and output multiple landmark specific feature descriptors by using image specific weighted average pooling on the feature maps, finally, the *Compare* module compares these landmark specific feature vectors between the two templates, and aggregates into *one* vector for final similarity prediction. The DCN is trained end-to-end for the task of template verification. The network is described in detail in § 3.

As a second contribution we introduce an idea from the instance retrieval literature to face template verification. Large scale instance retrieval systems achieved superior results by proceeding in two stages: given a query image, images are first retrieved and ranked using a very efficient method, such as bag of visual words; then, in a second stage, the top $k$ images are re-ranked using a more expensive method, such as geometric consistency with the query [15, 16]. Since image classification models can be trained very efficiently nowadays, we repurpose this re-ranking idea for template verification as follows: during training, we employ a standard image-wise classification model for sampling the hard template pairs. This is described in § 4, together with other training details, such as the training set and loss functions. In § 5, we report the verification performance of the DCN on the challenging IARPA Janus face recognition benchmarks – IJB-B [13] and IJB-C [14]. In both datasets, the DCN is able to *substantially* outperform the previous state-of-the-art methods.

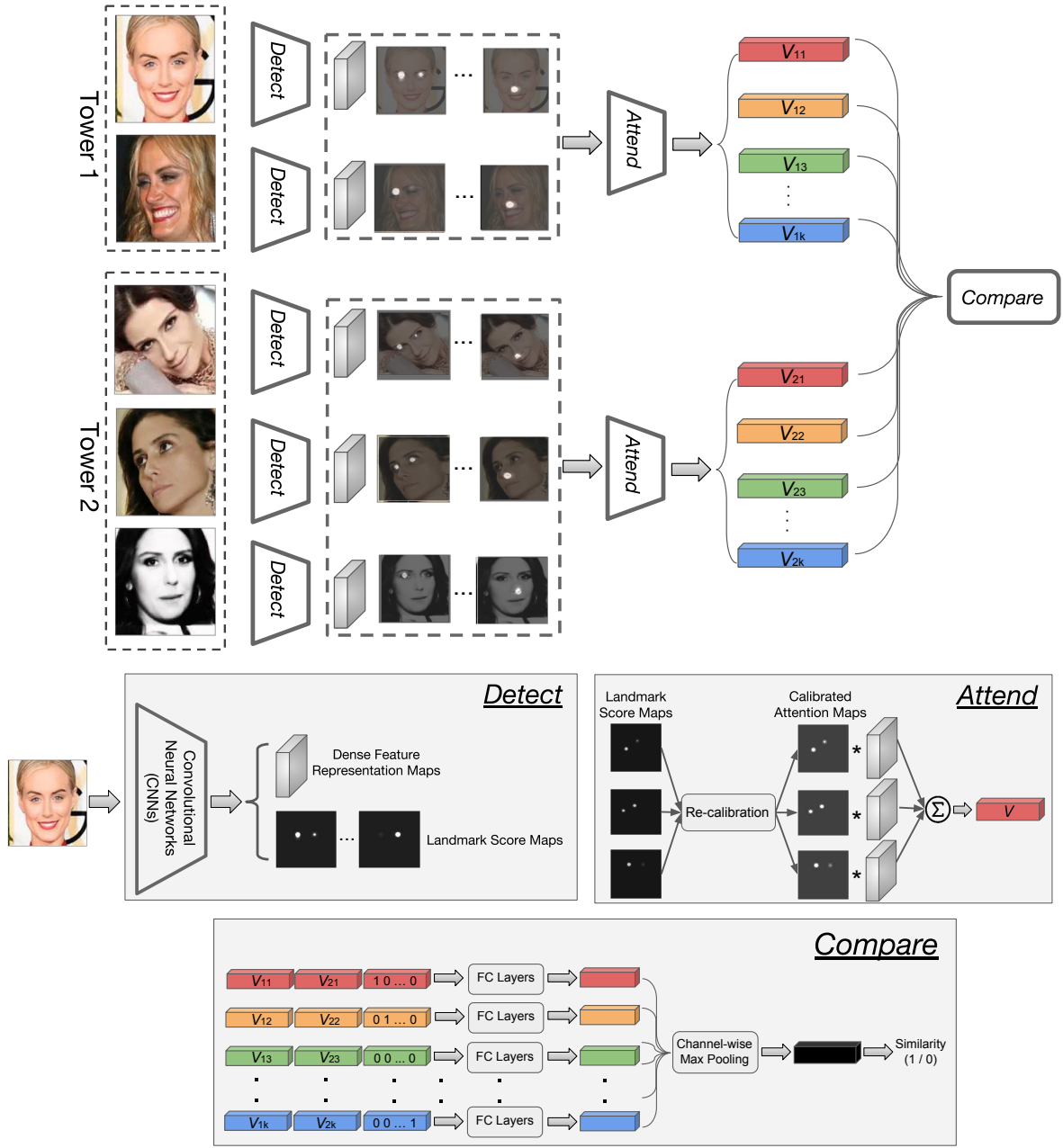

**Fig. 1.** *Top*: overview of the Deep Comparator Network (DCN)
*Bottom*: functionality of the individual modules, namely, *Detect*, *Attend*, *Compare*.
Each of the two towers in the DCN, is able to take a template (with an arbitrary number of images) as input. Each image is fed into the shared *Detect* module and outputs a feature map, as well as multiple discriminative landmark score maps. In the *Attend* module, landmark score maps (predicted from the same filter on different input images) are first re-calibrated within the template, then landmark specific feature vectors for each template are obtained by weighted average pooling on the feature maps. In the *Compare* module, landmark specific feature vectors between the two templates are compared with local "experts" (parametrized as fully connected layers), and aggregated into *one* vector for final similarity prediction.

## 2    Related work

In this section we review the work that has influenced the design of the DCN.

**Multi-column architectures.** Recent works [17–19] extend the traditional image-wise architectures to multi-columns, where the models are designed to take a set of images as inputs, and produce a single vector representation for the entire template. The model is trained to fuse useful information from multiple inputs by a weighting based on the image "quality"; for instance, high-resolution, frontal faces are weighted more than those under extreme imaging conditions or poses. However, these models still encode the entire template with *one* vector. They cannot tackle the challenge of *local landmark comparison* and *between template weightings*.

**Face recognition based on part representations.** Several previous works proposed to use part-based representation for a the face image or tacks. In [20], the face image is densely partitioned into overlapping patches at multiple scales, and each of the patches is represented by local features, such as Local Binary Pattern (LBP) or SIFT, then represented as a bag of spatial-appearance features by clustering. In [21], the Fisher Vector (FV) encoding is used to aggregate local features across different video frames to form a video-level representation.

**Attention models.** Attention models have been successfully used in machine translation [22], multiple object recognition [23], and image captioning [24]. In [25], the authors propose to extract part-based feature representations from a single input image with attention, and perform fine-grained classifications with these part specific representations. In general, the idea of these attentional pooling can be seen as a generalization of average or max pooling, where the spatial weights are obtained from a parametrized function (usually a small neural network) mapping from input image to an attention mask. Apart from soft attention, [26] proposed the Spatial Transformer Networks (STNs) that allows to learn whichever transformation parameters best aid the classification task. Although no ground truth transformation is specified during training, the model is able to attend and focus on the object of interest.

**Relation/co-occurrence learning.** In [27], in order to perform spatial relational reasoning, the features at every spatial location are modelled with the features at every other location. To model the co-occurence statistics of features, e.g. "brown eyes", a bilinear CNN [28] was proposed for fine-grained classification problems, the descriptor of one image is obtained from the outer product of the feature maps. As for few-shot learning, in [29], the authors propose to learn a local similarity metric with a deep neural network. As an extension, [30] experiments with models with more capacity, where the feature maps of images (from a support set and test set) are concatenated and passed to a relation module for similarity learning. Similarly, in this paper, we parameterize local "experts" to compare the feature descriptors from two templates.

## 3     Deep Comparator Networks

We consider the task of template-based verification, where the objective is to decide if two given templates are of the same object or not. Generally, in verification problems, the label spaces of the training set and testing set are disjoint. In the application considered here, the images are of faces, and the objective is to verify whether two templates show the same person or not.

From a high-level viewpoint, Deep Comparator Network (DCN) focus on the scenario that two templates (each has an arbitrary number of images) are taken as inputs, and trained end-to-end for template verification (as shown in Figure 1). We first overview the function of these modules: *Detect*, *Attend* and *Compare*, then give more details of their implementation. The detailed architectures for the individual modules are given in the **Supplementary Material**.

The *Detect* module is shared for each input image, dense feature maps and attention maps for multiple discriminative parts are generated for each image. In the face recognition literature, these discriminative parts are usually termed "landmarks", we will use this term from here on. Note that, the implicitly inferred landmarks aim to best assist the subsequent template verification task, they may not follow the same intuitions as human defined facial landmarks, e.g. mouth, nose, etc. Ideally, given a template with multiple images in various poses or illuminations, the landmark filters can be sensitive to different facial parts, viewpoints, or illuminations, e.g. one may be sensitive to the eyes in a frontal face, one may be more responsive to a mouth in a profile face. The *Detect* module acts as the base for fulfilling template comparison conditioned on *viewpoints/local landmarks.*

The *Attend* module achieves the *within template weighting* with an internal competition mechanism, and pools out multiple landmark-specific feature descriptors from each template. Given a template with multiple images, we hope to emphasize the feature representations from the relatively high quality images, while suppressing the lower ones. To achieve this, we normalize the attention score maps (inferred from different samples with the same landmark filter) into a probability distribution Consequently, multiple landmark specific feature descriptors are calculated by attending to the feature maps with image specific attentional masks (and it is assumed that high quality images will score more highly than aberrant ones, e.g. blurry images). Therefore, the contribution of aberrant images is suppressed, and viewpoint factors and facial parts are decomposed and template-wise aligned.

Finally, we use the *Compare* module to achieve the *between template weighting*. The template-wise verification is reformulated as the comparison conditioned on both global and local regions (i.e. landmarks), votings from the local "experts" are aggregated into *one* vector for the final similarity prediction.

### 3.1   Detect

The *Detect* module takes an image as input, and generates an intermediate dense representation with multiple ($K$) landmark score maps. Formally, we parametrize the module as a standard ResNet ($\psi(\cdot; \theta_1)$) shared by $n$ images (Figure 2 shows an example where $n = 3$):

$$[F_1, F_2, ..., F_n, A_1, A_2, ..., A_n] = [\psi(I_1; \theta_1), \psi(I_2; \theta_1), ..., \psi(I_n; \theta_1)] \qquad (1)$$

each input image is of size $I \in R^{W \times H \times 3}$, the output dense feature representation map $F \in R^{\frac{W}{8} \times \frac{H}{8} \times C}$, and a *set* of attention maps $A \in R^{\frac{W}{8} \times \frac{H}{8} \times K}$, where $W, H, C, K$ refer to the width, height, channels, and the number of landmark score maps respectively. A global score map is also obtained by a max over the local landmark score maps.

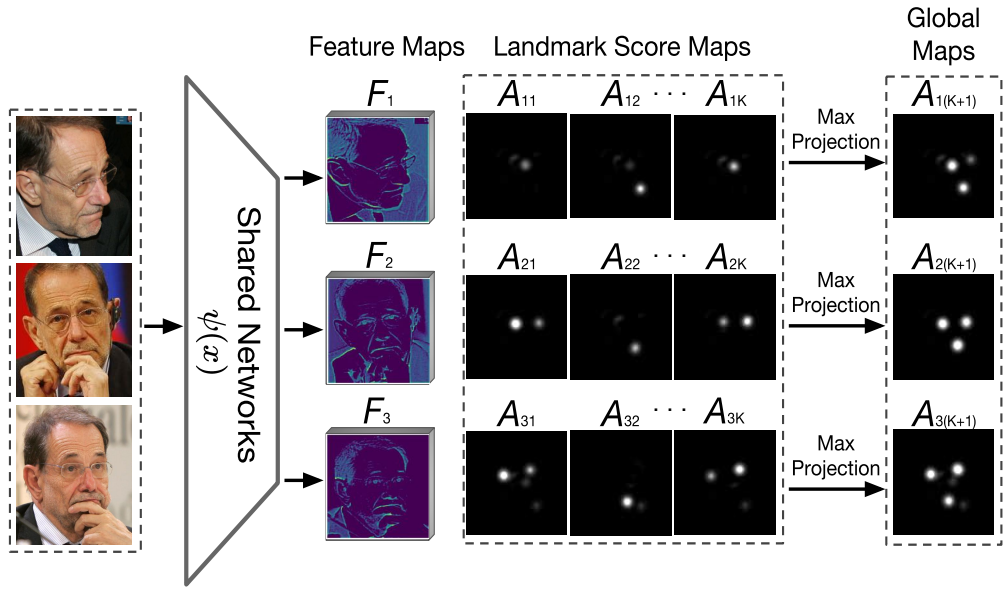

**Fig. 2.** The detect module. For each input image the detect module generates an intermediate feature map ($F's$), $K$ landmark attention maps ($A's$), and a global map (obtained by applying a max on the $A's$ channel dimension). In this example there are three input images and three of the $K$ landmark attention maps are shown. .

Ideally, the local score maps for each image should satisfy two conditions, *first*, they should be mutually exclusive (i.e. at each spatial location only one landmark is activated); *second*, the scores on the maps should positively correlate with image quality, i.e. the response of a particular landmark filter should be higher on high-resolution frontal images than on low-resolution frontal images.

### 3.2 Attend

*Re-calibration (Internal Competition).* Given the feature maps and landmark score maps for each input image, cross-normalization is used among the score maps within same template for *re-calibration*. Based on the "quality" of images within the template, the score maps (from different images within a single template) that localize same landmark are normalized as a distribution of weightings. Therefore, no matter how many images are packed into a template, the outputted attention maps in the same column always add up to 1.0 (Figure 1). Formally, for every $n \in [1, N]$ and $k \in [1, K]$:

$$A_{n..k} = \frac{exp(A_{n..k})}{\sum\limits_{nij} exp(A_{nijk})} \qquad (2)$$

*Attentional Pooling.* With the re-calibrated attention maps for each input image, we next attend to the spatial locations and compute local representations by the Hadamard Product. Formally, for each of the input image ($n \in N$), with the feature map as $F_n$, and one set of attention maps $A_n$,

$$V_k = \sum\limits_{nij} F_{nij:} \odot A_{nijk} \qquad \text{for } k \in [1 : K+1] \qquad (3)$$

Therefore, for each input template, we are able to calculate $K+1$ feature descriptors ($K$ landmark specific descriptors, "1" global feature descriptor), with each descriptor representing either one of the facial landmarks or global information.

### 3.3 Compare

Up to this point, we have described how to pool $K + 1$ feature vectors from the single template. In this module, we compare these descriptors in pairs between two different templates. In detail, the landmark-specific descriptors from two templates are first L2 normalized, and concatenated along with an one-hot encoded landmark identifier. Each concatenated vector is the input to a local "expert" parametrized by the fully connected (FC) layers [27]. Overall, the local experts are responsible for comparing the landmark-specific descriptors from different templates.

Formally, we learn a similarity function $y = C(x; \theta_2)$, where $x = [V_{1k} : V_{2k} : \text{ID}_{one-hot}]$, as shown in Figure 1. After passing through the fully connected layers, the feature representations given by local "experts" are max pooled, and fused to provide the final similarity score.

*Discussion.* Unlike the approaches of [28, 27], where features at every spatial location are compared with those at every other location, the compare module here only compares the descriptors that encode the same landmark region, e.g. frontal mouth to frontal mouth. By attaching the landmark identifier (the one-hot indicator vector), the fully connected layers are able to specialize for individual landmark. At a high level, leveraging multiple local experts in this way is similar to the use of multiple components (for different visual aspects) in the deformable part model (DPM) [31].

## 4   Experimental details

### 4.1   VGGFace2 Dataset

In this paper, all models are trained on the training set of large-scale VGGFace2 dataset [5], which has large variations in pose, age, illumination, ethnicity and profession (e.g. actors, athletes, politicians).

### 4.2   Landmark Regularizers

In the *Attend* module, the landmark score maps can be considered as a generalization of global average pooling, where the spatial "weights" are inferred implicitly based on the input image. However, in the *Detect* module, there is nothing to prevent the network from learning $K$ identical copies of the same landmark, for instance, it can learn to always predict the average pooling mask, or detect the eyes, or given a network with large enough receptive field, it can always pinpoint the centre of the image. To prevent this, we experiment with *two* different types of landmark regularizers: a diversity regularizer (unsupervised learning) and a keypoints regularizer (supervised learning).

*Diversity Regularizer [32].* In order to encourage landmark diversity, the most obvious approach is to penalize the mutual overlap between the score maps of different landmarks. Each of the landmark score maps is first self-normalized into a probability distribution ($p$'s) by using the softmax (Eq 4),

$$p_{nijk} = \frac{exp(A_{nijk})}{\sum_{ij} exp(A_{nijk})} \tag{4}$$

where $n, i, j, k$ refer to the image index within the template, width, height, number of attention maps respectively.

Ideally, if all $K$ landmarks are disjoint from each other, by taking the max projection of these normalized distribution, there should be exactly $K$ landmarks, and they should sum to $K$.

$$\mathcal{L}_{reg} = nK - \sum_{nij} \max_{k=1,..,K} p_{nijk} \tag{5}$$

Note here, this regularizer is zero only if the activations in the different normalized landmark score maps are disjoint and exactly 1.0.

*Keypoints Regularizer.* Benifiting from the previous fruitful research in facial keypoint detection, we obtain pseudo groundtruth landmarks from pre-trained keypoint detectors. Although the predictions are not perfect, we conjecture that they are sufficiently accurate to guide the network training at the early stages, and, as the training progresses, the regularizer weights is scheduled to decay, gradually releasing the parameter search space. As preprocessing, we predict 5 facial keypoints (Figure 3) over the entire dataset with a pre-trained MTCNN [33], and estimate three face poses by thresholding angle ratios.[1]

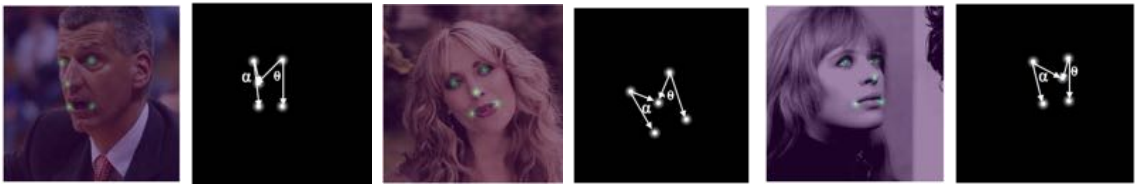

**Fig. 3.** Facial landmark detection for VGGFace2 images.
Face poses are quantized into three categories based on the ration $\alpha/\theta$. Left-facing profile : $\alpha/\theta < 0.3$, right-facing profile: $\alpha/\theta > 3.0$, frontal face: $\alpha/\theta \in [0.3, 3.0]$

Similar to the diveristy regularizer, the inferred landmark score maps are also self-normalized first (Eq 4), $\mathcal{L}2$ loss between the prediction ($p$) and the pseudo groundtruth ($\hat{p}$) is applied as auxiliary supervision. Note that, given each face image belongs to only one of the three poses, only 4 of the 12 landmark maps are actually useful for supervising an individual image.

$$\mathcal{L}_{reg} = \begin{cases} \sum_{nij} \frac{1}{2}(p_{nijk} - \hat{p}_{nijk})^2 \text{ for k in \{pose-specific keypoints\}} \\ \\ 0 \end{cases} \tag{6}$$

To make the experiments comparable, in both experiments, we use $K = 12$ landmark score maps in the *Detect* module.

### 4.3   Loss Functions

The proposed comparator network is trained end-to-end by optimizing three types of losses simultaneously: *first*, template-level identity classification loss, using a global feature representation obtained by attentional pooling with the re-calibrated global maps (refer to Figure 2); *second*, a standard classification loss (2 classes) on the similarity prediction from the *Compare* module; *third*, a regularization loss from the landmark score maps in the *Detect* module.

$$\mathcal{L} = \alpha_1(\mathcal{L}_{cls1} + \mathcal{L}_{cls2}) + \alpha_2\mathcal{L}_{sim} + \alpha_3\mathcal{L}_{reg} \tag{7}$$

---

[1] In our training, we only use 4 facial landmarks, left-eye, right-eye, nose, mouth. The mouth landmarks are obtained by averaging the two landmarks at mouth corners.

where $\alpha_1 = 2.0, \alpha_2 = 5.0$ refer to the loss weights for classification and similarity prediction, $\alpha_3$ refers to the weights for regularizer, which was initialized as 30.0 and decayed by half every $60,000$ iterations. Note that, $\alpha_3$ is scheduled to decrease, thus, even for the training with the keypoints regularizer, the auxiliary supervision only guides the network training in early stages. Thereafter, the classification and verification loss will dominate the training of these landmark localizations.

### 4.4    Hard-sample Mining

In order to train the Comparator Network for re-ranking, we need a method to sample hard template-template pairs. Here we described the procedure for this. The key idea is to use the features generated by a standard ResNet-50 trained for face image classification (on the VGGFace2 training set) to approximate the template descriptor, and use this approximate template descriptor to select hard template pairs.

In detail, the template-level descriptors are obtained by averaging the feature vectors (pre-computed from ResNet-50) of 3 images and L2-normalized. The selection of hard template pairs is then integrated into the training of the comparator network. At each iteration 256 identities are randomly sampled, and used to create 512 templates with 3 images in each template (i.e. two templates for each identity). In total, there are 256 positive template pairs, and a large number of negative pairs.

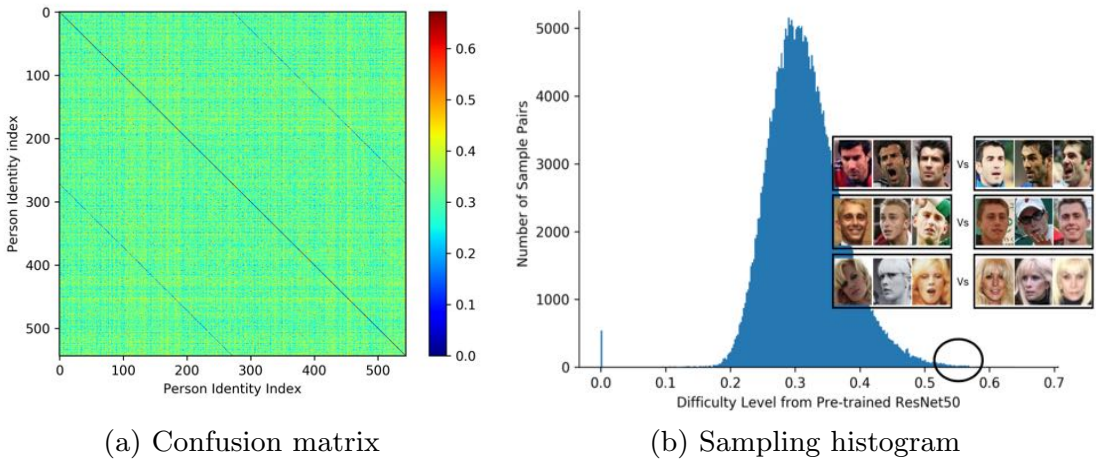

(a) Confusion matrix          (b) Sampling histogram

**Fig. 4.** Sampling strategy based on the pre-trained single-image classification networks. Larger values refer to more difficult template pairs.

By calculating the cosine similarity between different pairs of templates, we generate a $512 \times 512$ similarity matrix $M_s$ for the template-to-template verification, where small values refer to the predicted dissimilar pairs from the pre-trained ResNet50. We further define the verification difficulty matrix as:

$$d = |groundtruth - M_s| \qquad (8)$$

where groundtruth label is either 0 (dissimilar) or 1 (similar). Therefore, in the difficulty matrix, small values refer to the easy sample pairs, and large values refer to the difficult samples.

### 4.5   Training details

We train the entire Comparator Network end-to-end from scratch on the VG-GFace2 dataset, detailed architecture description can be found in **Supplementary Material**. During training, the shorter side of the input image is resized to 144, while the long side is center cropped, making the input images $144 \times 144$ pixels with the face centered, and 127.5 is subtracted from each channel. In each tower, 3 images are packed as a template input. Note that, there is a probability of 20% that the 3 images within one template are identical images[2]. In this case, the Comparator Network become equivalent to training on single image. Data augmentation is operated separately for each image with probability of 20%, including flipping, gaussian blur, motion blur, monochrome transformation. Adam [34] is used for optimization with an initial learning rate of $1e^{-4}$, and mini-bateches of size 64, with equal number of positive and negative pairs. The learning rate is decreased twice with a factor of 10 when errors plateau. Note that, although the batch size is small, the network is actually seeing $64 \times 6$ images every training step. Also, although the network is only trained with 3 images per tower, at test time it can be applied to any number of images per template.

Note, an alternative is to use a pre-trained face network, e.g. a ResNet-50 [5], as this considerably accelerates the training, compared to end-to-end training, with almost negligible loss in performance. In detail, the *Detect* module is replaced with the pre-trained ResNet-50; the landmark-conditioned descriptors are computed from the last layer of the *conv3* block (1/8 of the input spatial resolution); and the global descriptor comes from the last layer (average pooled vector).

## 5   Results

We evaluate all models on the challenging IARPA Janus Benchmarks, where all images are captured from unconstrained environments and show large variations in viewpoints and image quality. In contrast to the traditional closed-world classification tasks [1–3], verification is an open-world problem (i.e. the label spaces of the training and test set are disjoint), and thus challenges the capacity and generalization of the feature representations. All the models are evaluated on the standard 1:1 verification protocol (matching between the Mixed Media probes and two galleries), the performance is reported as the true accept rates (TAR) vs. false positive rates (FAR) (i.e. receiver operating characteristics (ROC) curve).

---

[2] This guarantees a probability of 64% that both templates contain 3 different images, and a probability of 36% that at least one template contains 3 identical image.

*IJB-B Dataset [13]* The IJB-B dataset is an extension of IJB-A [35], having 1,845 subjects with 21.8K still images (including 11,754 face and 10,044 non-face) and 55K frames from 7,011 videos.

| Model | 1:1 Verification TAR | | | |
|---|---|---|---|---|
| | FAR=1E − 4 | FAR=1E − 3 | FAR=1E − 2 | FAR=1E − 1 |
| Whitelam *et al.* [13] | 0.540 | 0.700 | 0.840 | −− |
| Navaneeth *et al.* [36] | 0.685 | 0.830 | 0.925 | 0.978 |
| ResNet50 [5] | 0.784 | 0.878 | 0.938 | 0.975 |
| SENet50 [5] | 0.800 | 0.888 | 0.949 | 0.984 |
| ResNet50+SENet50 | 0.800 | 0.887 | 0.946 | 0.981 |
| MN-v [19] | 0.818 | 0.902 | 0.955 | 0.984 |
| MN-vc [19] | 0.831 | 0.909 | 0.958 | 0.985 |
| ResNet50+DCN(Kpts) | **0.850** | 0.927 | 0.970 | 0.992 |
| ResNet50+DCN(Divs) | 0.841 | 0.930 | 0.972 | 0.995 |
| SENet50+DCN(Kpts) | 0.846 | 0.935 | 0.974 | 0.997 |
| SENet50+DCN(Divs) | 0.849 | **0.937** | **0.975** | **0.997** |

**Table 1.** Evaluation on 1:1 verification protocol on IJB-B dataset. (Higher is better) Note that the result of Navaneeth *et al.* [36] was on the Janus CS3 dataset.
DCN(Divs) : Deep Comparator Network trained with Diversity Regularizer
DCN(Kpts): Deep Comparator Network trained with Keypoints Regularizer.

*IJB-C Dataset [14]* The IJB-C dataset is a further extension of IJB-B, having 3,531 subjects with 31.3K still images and 117.5K frames from 11,779 videos. In total, there are 23124 templates with 19557 genuine matches and 15639K impostor matches.

| Model | 1:1 Verification TAR | | | |
|---|---|---|---|---|
| | FAR=1E − 4 | FAR=1E − 3 | FAR=1E − 2 | FAR=1E − 1 |
| GOTS-1 [14] | 0.160 | 0.320 | 0.620 | 0.800 |
| FaceNet [14] | 0.490 | 0.660 | 0.820 | 0.920 |
| VGG-CNN [14] | 0.600 | 0.750 | 0.860 | 0.950 |
| ResNet50 [5] | 0.825 | 0.900 | 0.950 | 0.980 |
| SENet50 [5] | 0.840 | 0.910 | 0.960 | 0.987 |
| ResNet50+SENet50 [5] | 0.841 | 0.909 | 0.957 | 0.985 |
| MN-v [19] | 0.852 | 0.920 | 0.965 | 0.988 |
| MN-vc [19] | 0.862 | 0.927 | 0.968 | 0.989 |
| ResNet50+DCN(Kpts) | 0.867 | 0.940 | 0.979 | 0.997 |
| ResNet50+DCN(Divs) | 0.880 | 0.944 | 0.981 | 0.998 |
| SENet50+DCN(Kpts) | 0.874 | 0.944 | 0.981 | 0.998 |
| SENet50+DCN(Divs) | **0.885** | **0.947** | **0.983** | **0.998** |

**Table 2.** Evaluation on 1:1 verification protocol on IJB-C dataset. (Higher is better) Results of GOTS-1, FaceNet, VGG-CNN are read from ROC curve in [14].

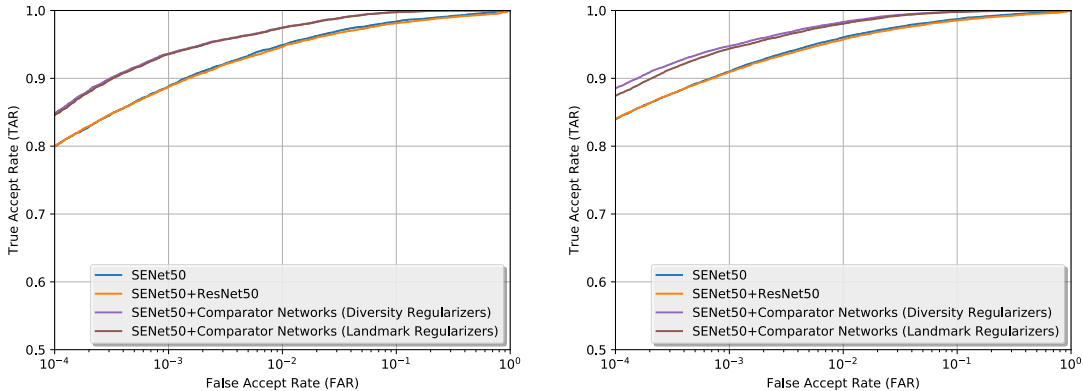

(a) ROC for IJB-B (Higher is better)       (b) ROC for IJB-C (Higher is better)

**Fig. 5.** ROC curve of 1:1 verification protocol on IJB-B & IJB-C dataset.

### 5.1   Discussion

Three phenomena can be observed from the evaluation results: *first*, comparing with the previous state-of-the-art model [5], the DCN trained by re-ranking can boost the performance significantly on both IJBB and IJBC (about $4-5\%$, which is a substantial reduction in the error); *second*, although the ResNet50 and SENet50 are designed differently and trained separately, ensembles of them do not provide any benifit. This shows that the difficult template pairs for ResNet50 remains difficult for another more powerful SENet50, indicating that the different models trained on single image classification are not complementary to each other; while in contrast, the DCN can be used together with either ResNet50 or SENet50 to improve the recognition system; *third*, the performance of DCN trained with different regularizers are comparable to each other, showing that groundtruth of facial keypoints is not critical in training DCN.

### 5.2   Visualization

Figure 6 shows the attention maps for a randomly sampled template that contains multiple images with varying poses. Visualizing the maps in this way makes the models interpretable, as it can be seen what the landmark detectors are concentrating on when making the verification decision. The *Detect* module has learnt to pinpoint the landmarks in different poses consistently, and is even tolerant to some out-of-plane rotation. Interestingly, the landmark detector actually learns to localize the two eyes simultaneously; we conjecture, that this is due to the fact that human faces are approximately symmetric, and also during training, the data is augmented with horizontal flippings.

## 6   Conclusion

We have introduced a new network that is able to compare templates of images and verify if they match or not. The network is very *flexible*, in that the number of images in each template can be varied at test time, it is also *opportunistic* in that it can take advantage of local evidence at test time, such as a specific facial features like a tattoo or a port-wine stain that might be lost in a traditional single tower per face encoding. Its performance substantially improves the state-of-the-art on the recent and very challenging IJB benchmarks.

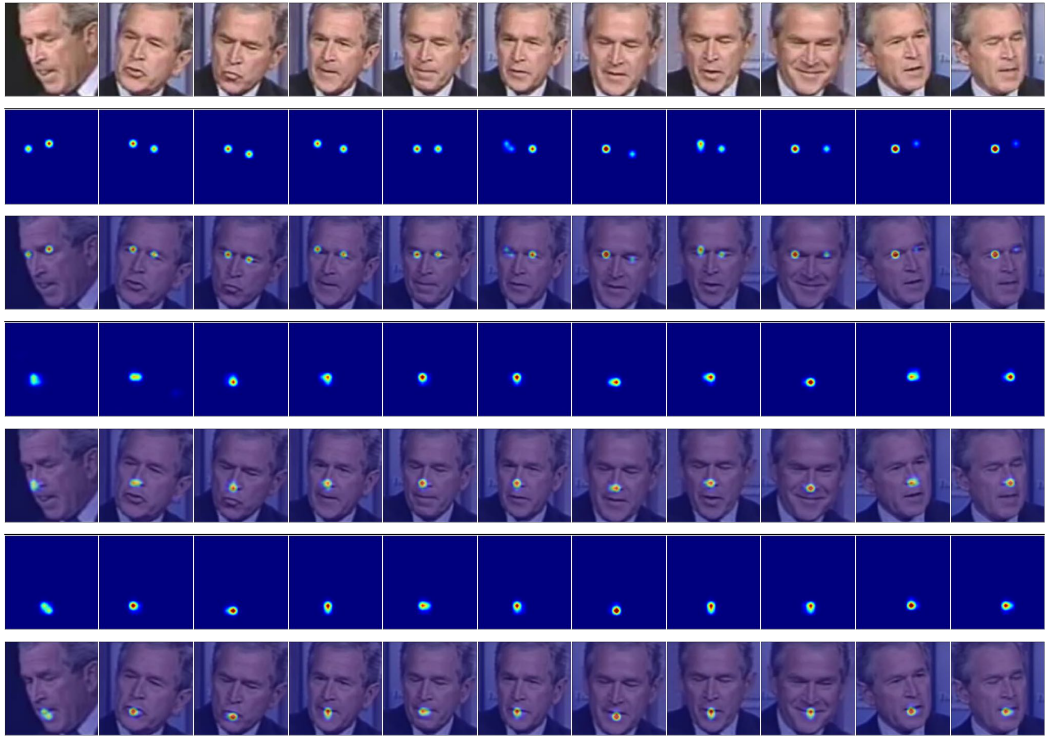

**Fig. 6.** Predicted facial landmark score maps after self-normalizing for three of the landmark detectors. Additional examples are given in the supplementary material. *1st row*: raw images in the template, faces in a variaty of poses are shown from left to right; *2nd,4th,6th row*: self-normalized landmark score maps (attention maps); *3rd, 5th, 7th row*: images overlayed with the attention maps.

Although we have used face templates in this work, the Comparator Network could be applied directly to person re-id, where often sets are available, and also potentially could be applied to other fine-grained classification tasks, e.g. to determine the species of a bird or flower from multiple images of the same instance.

## Acknowledgment

This research is based upon work supported by the Office of the Director of National Intelligence (ODNI), Intelligence Advanced Research Projects Activity (IARPA), via contract number 2014-14071600010. The views and conclusions contained herein are those of the authors and should not be interpreted as necessarily representing the official policies or endorsements, either expressed or implied, of ODNI, IARPA, or the U.S. Government. The U.S. Government is authorized to reproduce and distribute reprints for Governmental purpose notwithstanding any copyright annotation thereon.

# Supplementary Material

# S1   Comparator Network Architecture

In this section, we give the architectural details of the Comparator Network, covering the *Detect*, *Attend*, and *Compare* modules. As illustrated in the main paper, the Comparator Network takes two templates as input, and for training each template contains 3 images ($N = 3$) of the same person.

## S1.1   Detect

In the *Detect* module, a standard ResNet is shared for every input image, and outputs feature maps and multiple landmark score maps. In the paper, we use $K = 12$ local landmarks. The global map is obtained as the max projection of the local landmark maps.

| Module | Output Size (For each template) | Template 1 ($N \times 144 \times 144 \times 3$) | Template 2 ($N \times 144 \times 144 \times 3$) |
|---|---|---|---|
| *Detect* | $N \times 72 \times 72 \times 64$ | conv, $7 \times 7$, 64, stride 2 | |
| | $N \times 36 \times 36 \times 256$ | max pool, $3 \times 3$, stride 2 | |
| | | $\begin{bmatrix} \text{conv}, 1 \times 1, 64 \\ \text{conv}, 3 \times 3, 64 \\ \text{conv}, 1 \times 1, 256 \end{bmatrix} \times 3$ | |
| | $N \times 18 \times 18 \times 512$ | $\begin{bmatrix} \text{conv}, 1 \times 1, 128 \\ \text{conv}, 3 \times 3, 128 \\ \text{conv}, 1 \times 1, 512 \end{bmatrix} \times 4$ | |
| | $N \times 18 \times 18 \times 1024$ <br> $N \times 18 \times 18 \times (K+1)$ <br> $K$ local landmark maps <br> 1 global map | $\begin{bmatrix} \text{conv}, 1 \times 1, 256 \\ \text{conv}, 3 \times 3, 256 \\ \text{conv}, 1 \times 1, 1024 \end{bmatrix} \times 1024, \ \begin{bmatrix} \text{conv}, 1 \times 1, K \end{bmatrix}$ | |

Table 1: *Detect* Module.

## S1.2 Attend

In the *Attend* module, the local landmark score maps from the same filter are first re-calibrated (cross normalization), and then are used to pool the landmark specific feature descriptors. The normalization of the landmark score maps serves two important roles: first, low-quality images are effectively weighted down, therefore accomplishing the within-template weighting; second, it also makes it possible to ingest the template with an arbitrary number of images.

| Module | Output Size (For each template) | Template 1 $N \times 18 \times 18 \times 1024$ $N \times 18 \times 18 \times (K+1)$ | Template 2 $N \times 18 \times 18 \times 1024$ $N \times 18 \times 18 \times (K+1)$ |
|---|---|---|---|
| *Attend* | $N \times 18 \times 18 \times 1024$ $N \times 18 \times 18 \times (K+1)$ | Re-calibration | |
| | $(K+1) \times 1024$ | Attentional pooling | |
| | $(K+1) \times 1024$ | $L_2$ normalization | |

Table 2: *Attend* Module.

## S1.3 Compare

In the *Compare* module, the landmark specific feature descriptors for each template are compared using local "experts", that are parametrized as fully connected layers with 2048 nodes.

| Module | Output Size | Template 1 $(K+1) \times 1024$ | Template 2 $(K+1) \times 1024$ |
|---|---|---|---|
| *Compare* | $(K+1) \times (2048 + K + 1)$ | Concatenate template descriptors & part ID | |
| | $(K+1) \times 2048$ | Local experts (FC layer wiht 2048 nodes) | |
| | $1 \times 2048$ | Max pooling | |
| | $1 \times 2$ | Classification (1 or 0) | |

Table 3: *Compare* Module.

# S2 Sampling Strategy During Training

Hard template pairs are sampled from the CNNs pre-trained with standard face image classification. The top figure shows the estimated difficulty level distribution for template pairs from a pre-trained ResNet-50. Loosely speaking, they form a normal distribution with mean $\mu \in [0.3, 0.35]$. While training the Comparator Networks, we explicitly control the sampling process based on this distribution. As shown in the bottom figure, template pairs with difficulty level around $0.3, 0.5$ are used as training examples. Note, we do not focus on the extremly difficult pairs (above 0.6) for training, because they may arise from noisy labels.

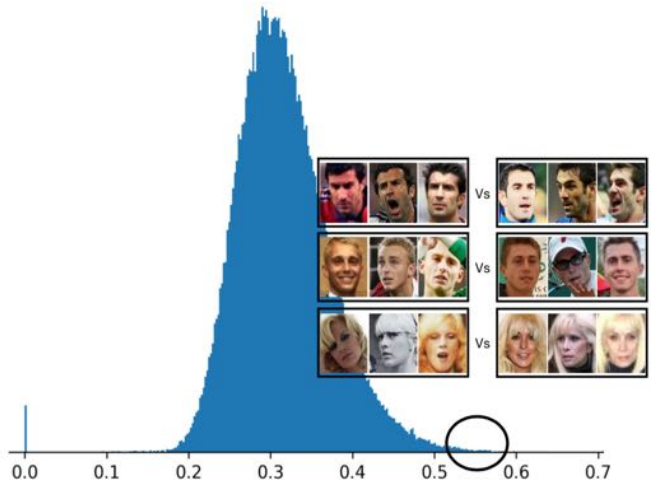

(a). Difficulty Level Distribution from the Pre-trained CNNs

Figure S1: Sampling strategy based on the pre-trained single-image classification networks. Larger values refer to more difficult template pairs. The Comparator Network is trained by sampling template pairs at different levels.

# S3    Other Network Architectures

| Output size | ResNet-50 | SENet-50 |
|---|---|---|
| $112 \times 112$ | conv, $7 \times 7$, 64, stride 2 | |
| $56 \times 56$ | max pool, $3 \times 3$, stride 2 | |
| | $\begin{bmatrix} \text{conv}, 1 \times 1, 64 \\ \text{conv}, 3 \times 3, 64 \\ \text{conv}, 1 \times 1, 256 \end{bmatrix} \times 3$ | $\begin{bmatrix} \text{conv}, 1 \times 1, 64 \\ \text{conv}, 3 \times 3, 64 \\ \text{conv}, 1 \times 1, 256 \\ fc, [16, 256] \end{bmatrix} \times 3$ |
| $28 \times 28$ | $\begin{bmatrix} \text{conv}, 1 \times 1, 128 \\ \text{conv}, 3 \times 3, 128 \\ \text{conv}, 1 \times 1, 512 \end{bmatrix} \times 4$ | $\begin{bmatrix} \text{conv}, 1 \times 1, 128 \\ \text{conv}, 3 \times 3, 128 \\ \text{conv}, 1 \times 1, 512 \\ fc, [32, 512] \end{bmatrix} \times 4$ |
| $14 \times 14$ | $\begin{bmatrix} \text{conv}, 1 \times 1, 256 \\ \text{conv}, 3 \times 3, 256 \\ \text{conv}, 1 \times 1, 1024 \end{bmatrix} \times 6$ | $\begin{bmatrix} \text{conv}, 1 \times 1, 256 \\ \text{conv}, 3 \times 3, 256 \\ \text{conv}, 1 \times 1, 1024 \\ fc, [64, 1024] \end{bmatrix} \times 6$ |
| $7 \times 7$ | $\begin{bmatrix} \text{conv}, 1 \times 1, 512 \\ \text{conv}, 3 \times 3, 512 \\ \text{conv}, 1 \times 1, 2048 \end{bmatrix} \times 3$ | $\begin{bmatrix} \text{conv}, 1 \times 1, 512 \\ \text{conv}, 3 \times 3, 512 \\ \text{conv}, 1 \times 1, 2048 \\ fc, [128, 2048] \end{bmatrix} \times 3$ |
| $1 \times 1$ | global average pool, 1000-d $fc$, softmax | |

Table 4: (**Left**) ResNet-50. (**Right**) SENet-50. The shapes and operations with specific parameter settings of a residual building block are listed inside the brackets and the number of stacked blocks in a stage is presented outside. The inner brackets following by $fc$ indicates the output dimension of the two fully connected layers in an SE module.

# S4 Visualization

## S4.1 Keypoints Regularizer

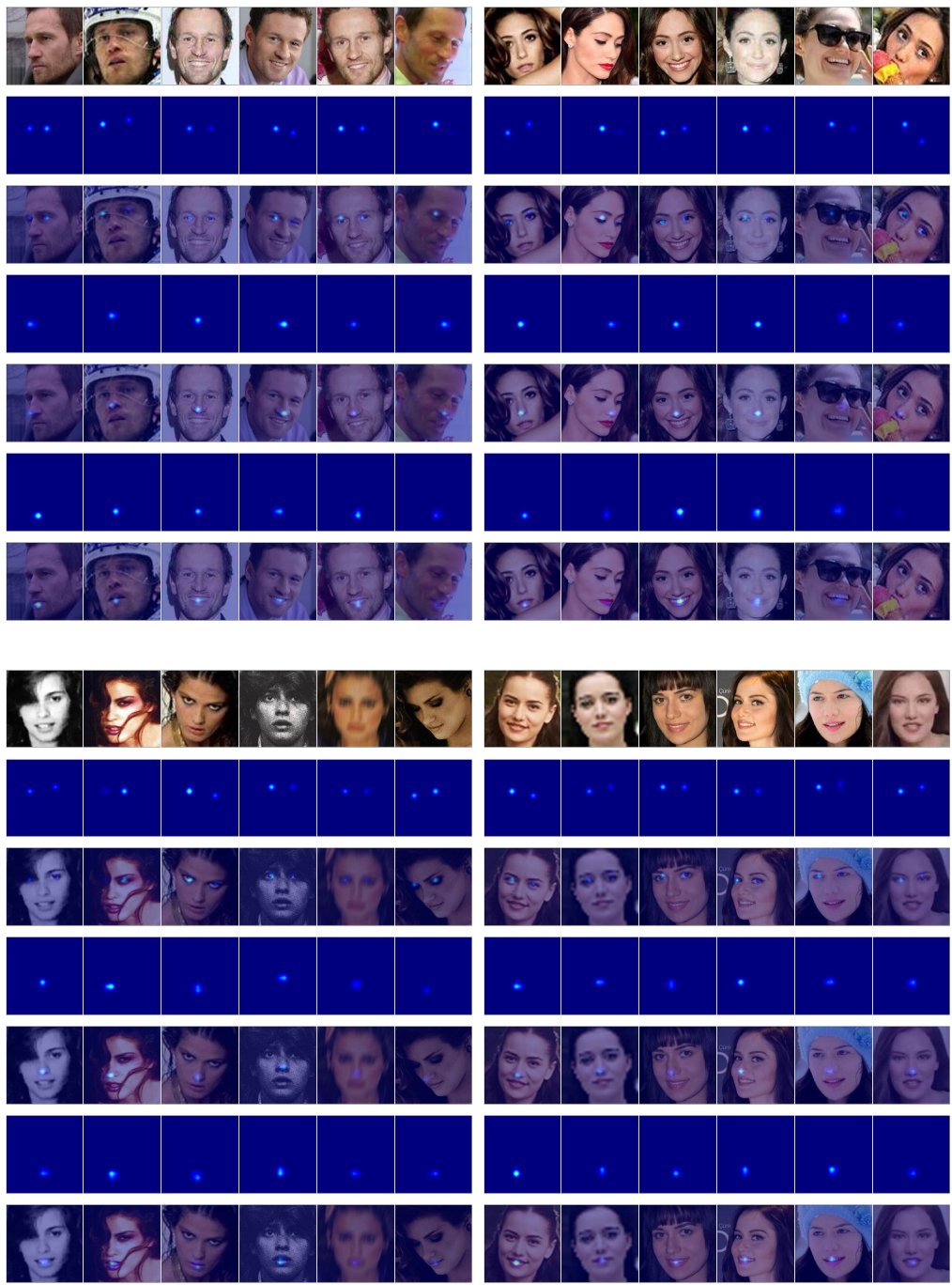

Figure S2: Predicted facial landmark score maps after self-normalizing for the landmark detectors. For each section, *1st row*: raw images in the template, faces in a variaty of poses are shown from left to right; *2nd,4th,6th row*: self-normalized landmark score maps (attention maps); *3rd, 5th, 7th row*: images overlayed with the attention maps.

## S4.2 Diversity Regularizer

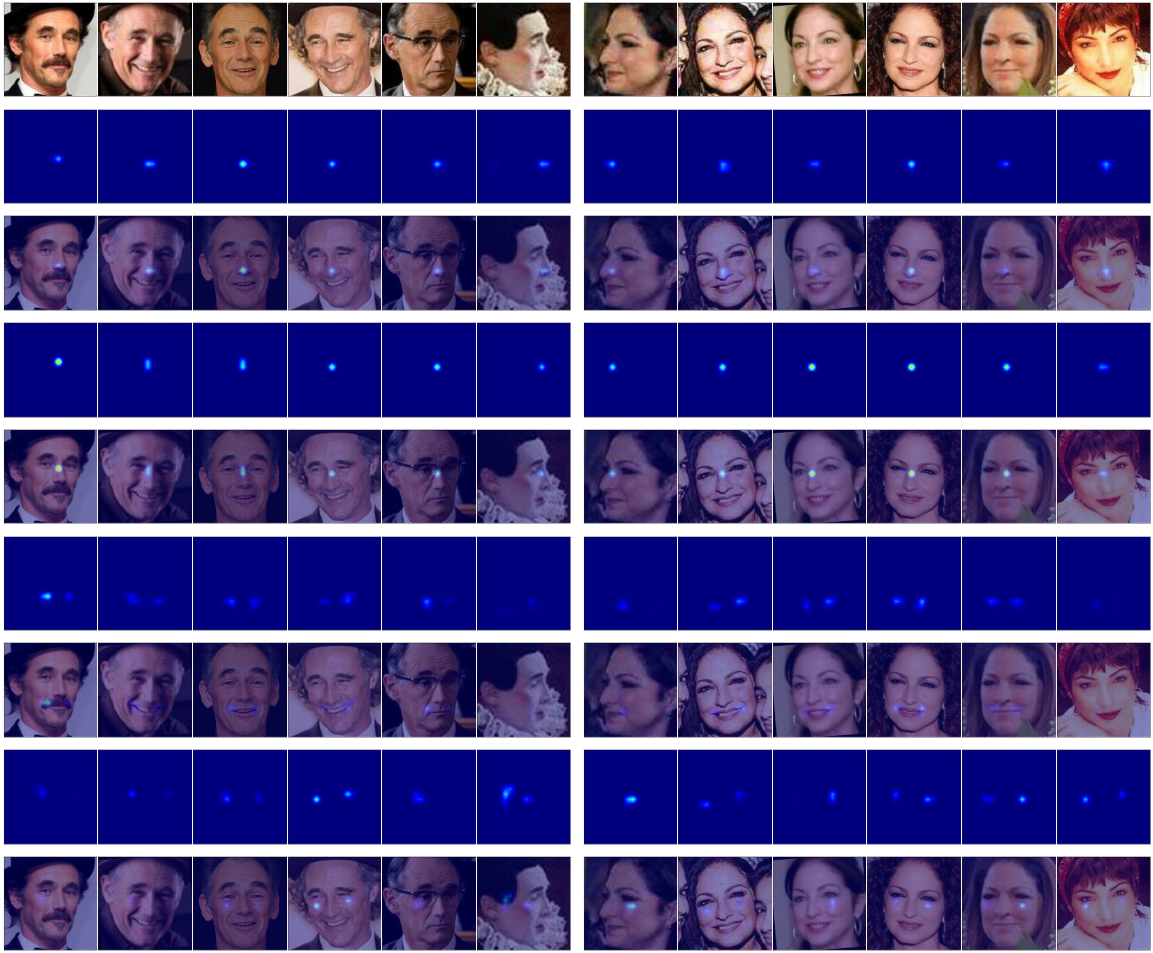

Figure S3: Predicted facial landmark score maps after self-normalizing for the landmark detectors. *1st row*: raw images in the template, faces in a variaty of poses are shown from left to right; *2nd,4th,6th, 8th row*: self-normalized landmark score maps (attention maps); *3rd, 5th, 7th, 9th row*: images overlayed with the attention maps.